\documentclass[conference]{IEEEtran}
\IEEEoverridecommandlockouts
\usepackage{cite}
\usepackage{amsmath,amssymb,amsfonts}
\usepackage{algorithmic}
\usepackage{graphicx}
\usepackage{textcomp}
\usepackage{xcolor}
\usepackage{lipsum}
\usepackage{tabularx}
\usepackage{caption}
\usepackage{geometry}
\def\BibTeX{{\rm B\kern-.05em{\sc i\kern-.025em b}\kern-.08em
    T\kern-.1667em\lower.7ex\hbox{E}\kern-.125emX}}
\begin{document}

\title{Reinforcement Learning-based Adaptive Path Selection for Programmable Networks}

\author{\IEEEauthorblockN{José E. Zerna Torres\textsuperscript{$*$}, Marios Avgeris\textsuperscript{$*$}, Chrysa Papagianni\textsuperscript{$*$}, \\Gergely
Pongrácz\textsuperscript{$*\S$}, István 
Gódor\textsuperscript{$\S$}, Paola
Grosso\textsuperscript{$*$}}

\IEEEauthorblockA{\textit{\textsuperscript{$*$}  Institute of Informatics, University of Amsterdam, Amsterdam, The Netherlands}  \\
\textit{\textsuperscript{$\S$} Ericsson Research, Budapest, Hungary} \\
\{j.e.zernatorres, m.avgeris, c.papagianni, p.grosso\}@uva.nl, \\
\{gergely.pongracz, istvan.godor\}@ericsson.com}

}

\date{}



\maketitle

\begin{abstract}
This work presents a proof-of-concept implementation of a distributed, in-network reinforcement learning (IN-RL) framework for adaptive path selection in programmable networks. By combining Stochastic Learning Automata (SLA) with real-time telemetry data collected via In-Band Network Telemetry (INT), the proposed system enables local, data-driven forwarding decisions that adapt dynamically to congestion conditions. The system is evaluated on a Mininet-based testbed using P4-programmable BMv2 switches, demonstrating how our SLA-based mechanism converges to effective path selections and adapts to shifting network conditions at line rate.
\end{abstract}


\section{Introduction}

Modern networks are increasingly dynamic and application-sensitive, driven by services such as cloud computing, real-time video, and Internet of Things (IoT). These trends are expected to intensify in the context of 6G, which targets immersive Extended Reality (XR), autonomous systems, and hyper-reliable low-latency communication (HRLLC) \cite{Tao2023}. To support such demands, networks must react to changing conditions in real time. Traditional rule-based mechanisms, while reliable, often lack the flexibility to handle changing conditions in real time \cite{Boutaba2018}. This has sparked growing interest in applying Machine Learning (ML) to networking systems to enable intelligent, adaptive control.

Software-Defined Networking (SDN) has played a key role in enabling programmable control over the network \cite{Feamster2014}. By decoupling the control and data planes, SDN architectures provide a centralized view of the network and enable the dynamic reprogramming of its behavior through external controllers. This has made it possible to apply ML techniques in the control plane, where models trained on historical data can support tasks such as traffic classification, anomaly detection, and load balancing. However, these approaches face significant limitations \cite{Xiong2019}, as they introduce control loop latency, rely on out-of-band telemetry, and therefore cannot respond effectively to rapid network events. As a result, they struggle to respond effectively to fine-grained changes in network state, particularly in latency-sensitive applications.

To overcome the limitations of centralized ML, recent efforts have explored \textit{in-network control}, where decisions are executed directly within programmable switches. The emergence of programmable data planes (PDP), enabled by languages like P4 \cite{Bosshart2014} and  platforms such as Tofino, SmartNICs, and FPGAs, makes it possible to embed the control logic directly in the network pipeline. This allows switches to react to local conditions at line rate, without involving the control plane. 

To enable systems to learn and adapt at runtime, Reinforcement Learning (RL) emerges as a natural fit. By learning through interaction with the environment, RL can adjust its behavior over time based on feedback signals such as latency, congestion, or link failures. Previous efforts have explored the use of RL in the programmable data plane, including tabular Q-learning approaches compiled to P4 or offloaded to SmartNICs \cite{Zheng2023, Simpson2022}. However, such methods rely on large state-action tables or gradient updates, which make them impractical for many data plane environments due to constraints in memory and arithmetic operations. This has created a demand for lightweight, feedback-driven learning algorithms that align with the limitations and abstractions of data-plane programming.

As a response, in this work, we present a lightweight, dataplane-native RL mechanism based on \textit{Stochastic Learning Automata (SLA)}. SLA is a model-free, low-memory RL algorithm that learns action preferences through probabilistic updates \cite{nsal1998}. Unlike traditional RL, SLA avoids explicit state representations and operates using only a compact action probability vector. The simplicity of SLA aligns well with the match-action abstraction of P4, making it a practical candidate for in-network learning under the constraints of programmable data planes. Our contribution is threefold:
\begin{enumerate}
    \item We model the path selection problem as a feedback-driven decision process, using two congestion metrics, queue length and dequeueing delay, to guide action updates in real time. This formalization enables distributed agents to evaluate path performance using only local observations and scalar rewards.
    \item We develop a fully dataplane-resident learning agent, implemented using the match-action pipeline of P4. The agent executes both learning and inference directly within the forwarding logic, leveraging in-network telemetry embedded in packets, to adapt path selection probabilities.
    \item We evaluate the proposed system on a programmable testbed in Mininet, demonstrating how our SLA-based mechanism converges to effective path selections and adapts to shifting network conditions with minimal overhead. 
\end{enumerate}

The remainder of this paper is organized as follows. Section \ref{sec:sota} reviews related work on in-network machine learning and SLA-based control. Section \ref{sec:sysmod} introduces the system model and formalizes the path selection problem. Section \ref{sec:inrl} introduces the RL-based algorithm for adaptive path selection, describing its learning dynamics and decision-making logic. In Section \ref{sec:evaluation} we describe the experimental setup and implementation decisions and the results of our experiments. Finally, Section \ref{sec:conclusions} concludes the paper and outlines potential directions for future research.

\section{Related Work}
\label{sec:sota}

The emergence of programmable data planes has sparked interest in deploying ML models directly within network devices to improve performance in terms of throughput and latency. Early work in this space focused on compiling classical supervised models—such as decision trees and support vector machines—into the data plane for tasks like packet classification. Frameworks like IIsy \cite{Zheng2022_IIsy} and Planter \cite{Zheng2024_Planter} demonstrated how such models could be mapped to match-action tables and embedded in targets like BMv2, Tofino, and NetFPGA. However, these systems rely on offline training and static model logic, limiting their ability to adapt to changing network conditions. This has motivated the exploration of in-network reinforcement learning (IN-RL) as a means to equip dataplane agents with online learning capabilities—enabling systems to dynamically adjust their behavior through direct interaction with the environment, and respond more efficiently to the demands of real-time, evolving network scenarios.

Initial efforts to bring RL into the dataplane are exemplified by OPaL \cite{Simpson2021}. OPaL leveraged idle processing units in SmartNICs to run tile-coded Sarsa inference at runtime. While this demonstrated the feasibility of local RL decision-making, OPaL's design is tightly coupled to the specific capabilities of the SmartNIC platform, limiting its portability and general applicability. More flexible approaches have since been explored. 

QCMP \cite{Zheng2023} presents the first complete realization of IN-RL on a switch ASIC, without the use of externs. Targeting the load balancing problem, QCMP employs Q-learning to update path weights in response to queue length telemetry. The system supports both control plane and dataplane-based Q-value updates, enabling a fully in-dataplane learning loop. However, its design remains closely tied to the tabular Q-learning formulation, which--while effective for small discrete environments--poses scalability challenges as the state-action space grows.

A recent and notable contribution is (QL)\textsuperscript{2}-RODAP. Polverini et al. in \cite{Polverini2025} propose a fully decentralized Q-learning-based routing mechanism implemented entirely in the data plane. Designed for delay-sensitive applications in 6G core networks, (QL)\textsuperscript{2}-RODAP selects per-packet forwarding paths based on queue occupancy, leveraging in-band telemetry for local state dissemination. Unlike prior approaches that separate training and inference, this work performs both phases in programmable switches, enabling fast reaction times and reducing end-to-end delay. However, (QL)\textsuperscript{2}-RODAP also relies on maintaining and updating distributed Q-tables, which can pose scalability and implementation challenges due to memory constraints and queue visibility limitations on real ASICs. In contrast, our work adopts a stateless learning approach based on SLA, avoiding Q-tables altogether and simplifying data plane integration.

Stochastic Learning Automata (SLA) \cite{nsal1998} offers a lightweight, model-free alternative. SLA maintains a compact action probability distribution, updated using scalar reward feedback, and avoids the storage and computation burdens of Q-learning or deep RL. Prior applications of SLA include VNF orchestration and wireless scheduling \cite{Avgeris2023}, but its potential for in-network decision-making remains unexplored. To the best of our knowledge, this is the first work to implement both training and inference of SLA entirely within the data plane, enabling adaptive, probabilistic path selection based on real-time telemetry, without control-plane involvement or external learning loops.

\section{System Model} 
\label{sec:sysmod}

We model our network infrastructure as a graph $\mathcal{G} = (\mathcal{N}, \mathcal{L})$, where $\mathcal{N}$ represents the set of network nodes, including both forwarding devices (switches) and end-hosts, and $\mathcal{L} \subseteq \mathcal{N} \times \mathcal{N}$ is the set of links that interconnect them. We define the set of hosts as $\mathcal{H} = \{h_1, h_2, \ldots, h_{|H|}\}\subseteq \mathcal{N}$ and a set of switches $\mathcal{S} = \{S_1, S_2, \ldots, S_{|S|}\}, \subseteq \mathcal{N}$, with $\mathcal{H} \cap \mathcal{S} = \varnothing$. Among the switches, we define a subset of decision nodes $\mathcal{Z}=\{z_1, \ldots, z_{|Z|}\}$,  where traffic is steered among alternative path segments, and a set of endpoint nodes $\mathcal{E} = \{e_1, e_2, \ldots, e_{|\mathcal{E}|}\}$, where the corresponding segments terminate. 

To structure the graph modularly, we define the concept of a domain. A domain $\mathcal{D}_k = (\mathcal{N}_k, \mathcal{L}_k) \subseteq \mathcal{G}$ is a connected subgraph that begins at a decision node $z_k \in \mathcal{Z}$ and ends at an endpoint node $e_k \in \mathcal{E}$. Here, the index $k \in \{1, \dots, |\mathcal{D}|\}$ identifies a specific domain. Each domain defines the scope within which a local forwarding decision at $z_k$ selects one among several alternative paths toward $e_k$. We denote the set of domains in the graph as $\{\mathcal{D}_1, \ldots, \mathcal{D}_{|\mathcal{D}|} \}$, where $|\mathcal{D}|$ is the total number of domains. Domains may overlap: a single node or link may belong to multiple domains, depending on the connectivity and the segmentation of paths across the graph.

\begin{figure}[t]
    \centering
    \includegraphics[width=\linewidth]{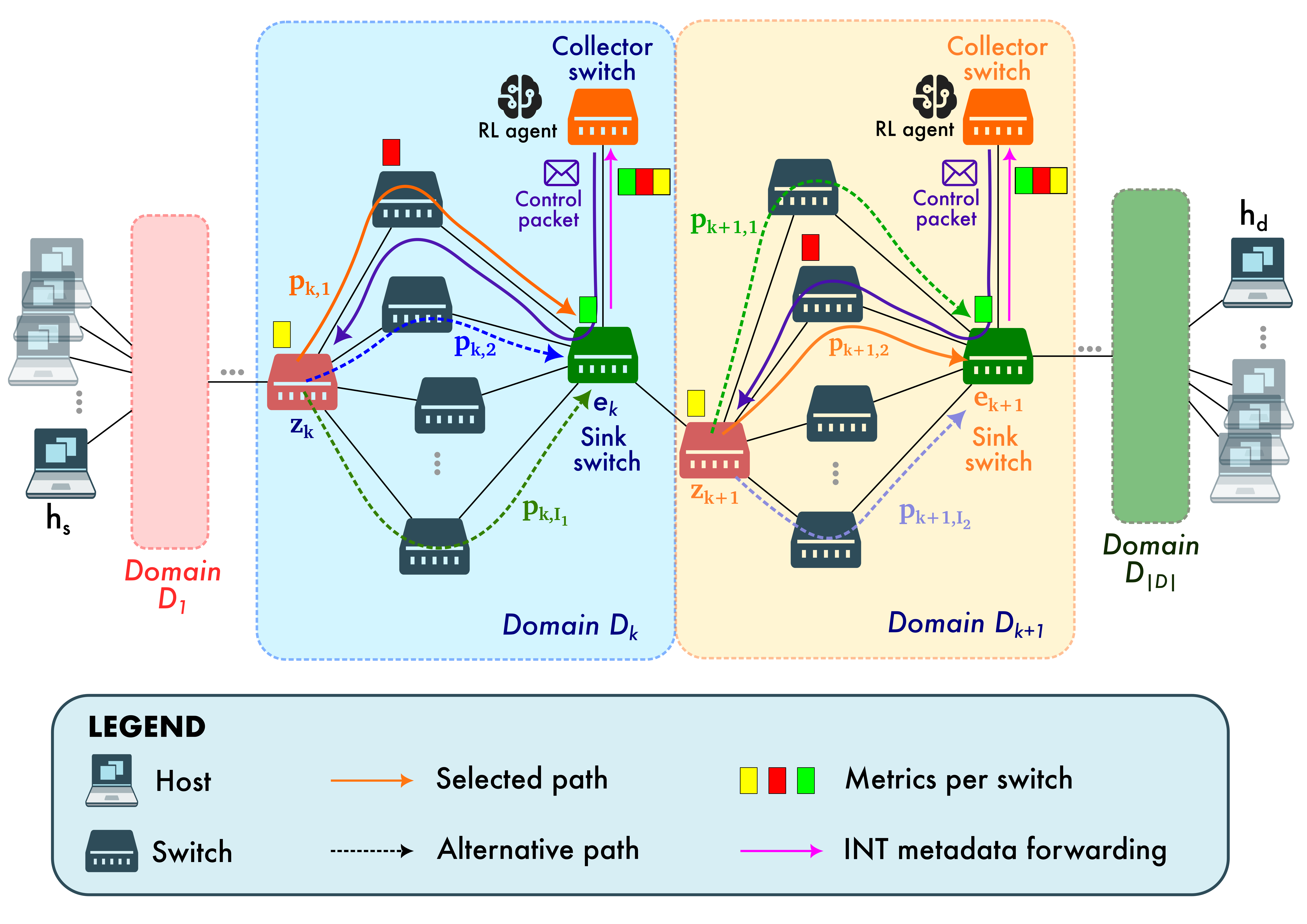}
    \caption{Data flow for in-network RL path selection.}
    \label{fig:general_topology}
\end{figure}

Among the hosts, we define a subset of source hosts $H_S \subseteq \mathcal{H}$ and a subset of destination hosts $H_D \subset \mathcal{H}$, involved in communication sessions. A session is defined by a communicating pair  $(h_s, h_d) \in H_S \times H_D$, where packets are transmitted from $h_s$ to $h_d$ through a sequence of forwarding switches. Here, $s$ and $d$ denote the indices of the specific source and destination hosts, respectively, involved in the communication. These paths traverse one or more domains, each beginning at a decision node and terminating at an endpoint node, with the forwarding behavior determined locally within each domain. The domain-based segmentation and network layout of our system are presented in Figure~\ref{fig:general_topology} .

For each domain $\mathcal{D}_k$, we define the set of feasible path segments as $\mathcal{P}_k = \{p_{k,1}, p_{k,2}, \ldots, p_{k,I_k}\}$, where $I_k$ is the number of candidate paths available in the domain $\mathcal{D}_k$. Each path segment $p_{k,i}$ is an ordered list of nodes in $\mathcal{N}$ that begins at the domain’s decision node $z_k$, ends at an associated endpoint node $e_k$, and traverses only switches in $\mathcal{D}_k$ in between. The index $i \in \{1, I_k\}$ denotes the available paths in the domain $\mathcal{D}_k$.

Each path segment $p_{k,i} \in \mathcal{P}k$ is evaluated using a set of network performance metrics $\mathcal{M}_{k,i} = \{m_{k,i}^{(1)}, m_{k,i}^{(2)}, \ldots, m_{k,i}^{(R)}\}$, where each ${m_{k,i}^{(r)}}$ represents the value of the $r$-th metric for path segment $p_{(k,i)}$. The selection and configuration of metrics can be tailored by the network operator or service provider based on the specific optimization objectives, offering flexibility to adapt the model to diverse network policies and service requirements. 

\section{IN-RL Adaptive Path Selection}
\label{sec:inrl}

We introduce an IN-RL adaptive path selection methodology that enables distributed, real-time traffic steering based on metrics collected using In-Band Network Telemetry (INT)~\cite{INTspecs}. In this approach, a RL agent is deployed within each domain $\mathcal{D}_k$ to evaluate the candidate path segments $\mathcal{P}_k$ and dynamically update forwarding decisions at the corresponding decision node $z_k$. The RL agent operates from a dedicated INT collector switch \cite{Alhamed2023}, where telemetry reports are received from the sink switch, and then aggregated and analyzed. Each agent computes its decisions using two key performance metrics $\mathcal{M}_{k,i}$ (i.e. $R=2$) related to congestion: \textit{queue length}, which reflects congestion along the path, and \textit{dequeueing delay}, which captures the time a packet waits in the queue before it is scheduled for transmission. These metrics jointly reflect congestion conditions along candidate paths, factors with direct impact on Quality of Service (QoS). While our current implementation focuses on these two metrics, the methodology is general and extensible, allowing additional or alternative indicators to be integrated based on operator objectives or application scenarios. The agent computes its decisions using two key performance metrics $\mathcal{M}_{k,i}$ (i.e. $R=2$) related to congestion: \textit{queue length}, which reflects congestion along the path, and \textit{dequeueing delay}, which captures the time a packet spends in the queue before transmission. These metrics were selected for their ability to jointly reflect congestion  conditions along candidate paths, factors that directly impact Quality of Service (QoS). While our current implementation focuses on these two metrics, the methodology is general and extensible, allowing additional or alternative indicators to be integrated based on operator objectives or application scenarios.


\subsection{In-Network RL Agent}
The \textit{In-Network RL Agent} is the central component of our adaptive path selection methodology. Implemented directly in the collector switch, the agent maintains a probability distribution over the candidate path segments $p_{(k,i)} \in \mathcal{P}_k$. These probabilities are updated iteratively based on the observed metrics from the selected segment path.

Upon receiving an INT report for a packet $t$, the agent evaluates the performance of the selected path segment $p_{(k,i)} \in \mathcal{P}_k$ using a reward function that incorporates the segment queue length ${Q}_{(k,i)}^{(t)}$ and the dequeueing delay ${D}_{(k,i)}^{(t)}$. The reward $\mathcal{R}^{(t)}_{(k,i)}$ is then computed as:
\begin{equation}
\mathcal{R}_{(k,i)}^{(t)} = \beta_1 \cdot f\left({Q}_{(k,i)}^{(t)}\right) + \beta_2 \cdot f\left({D}_{(k,i)}^{(t)}\right)
\label{eq:reward}
\end{equation}


where $\beta_1, \beta_2 \in [0,1]$ are weighting coefficients with \( \beta_1 + \beta_2 = 1 \), allowing the operator to tune the relative importance of each metric. To emphasize when a metric shows a performance degradation, we apply the sigmoid function $f$:
\begin{equation}
    f(m_{k,i}) =  1 - \frac{1}{1 + e^{-C(m_{k,i} - \tau_m)}}
    \label{eq:sigmoid_function}
\end{equation}
where $\tau_m$ is a target threshold defined by the network operator to reflect a QoS requirement for the metric $m_{k,i}$. The coefficient $C \in \mathbb{R}^{+}$ makes the function $f$ output close to 0 when the metric significantly exceeds the threshold and close to 1, otherwise. 

The SLA mechanism then updates the probabilities for the selected path and all the non-selected paths as follows:

\begin{align}
P_{k,i}^{(t+1)} &=
\begin{cases}
P_{k,i}^{(t)} + \alpha \mathcal{R}_{k,i}^{(t)} (1 - P_{k,i}^{(t)}), & \text{if $p_{k,i}$ is selected} \\
P_{k,i}^{(t)} - \alpha \mathcal{R}_{k,i^*}^{(t)} P_{k,i}^{(t)}, & \text{otherwise}
\end{cases}
\label{eq:sla_prob_update}
\end{align}

where \( \alpha \in (0,1) \) is the learning rate. This rule increases the probability of the selected path proportionally to the reward received, while decreasing the probabilities of the non-selected paths accordingly.

The agent then performs a probabilistic selection based on the current distribution, choosing the next packet's forwarding path segment. Once the selection is made, the agent communicates it to its corresponding decision node $z_k$ via a control packet, allowing the new path choice to be enforced in the data plane without centralized coordination.


To avoid oscillatory behavior and reduce unnecessary switching between paths, the RL agent operates in two phases. This phase-
based design allows the system to converge and stabilize once
a reliable path has been identified. During the \textit{Learning Phase}, forwarding decisions are made probabilistically, and rewards are computed based on the queue length and dequeueing delay observed from INT reports. Path selection probabilities are updated iteratively using the equation \ref{eq:sla_prob_update}. Convergence is achieved when one path's selection probability reaches a predefined threshold. Then the agent transitions to the \textit{Optimized Steering Phase}, where all traffic is forwarded along the learned path, effectively freezing the probability distribution to reduce unnecessary switching. 

To maintain awareness of changing network conditions, we have included a background probing mechanism, in which periodic packets are sent through the alternate path to monitor performance changes. The reward derived from these probe packets is used to monitor the alternative path without affecting the main forwarding behavior. If the observed performance of the selected path degrades or if the alternative path shows sustained improvement—detected through changes in the exponential moving average (EMA) of the reward— the agent triggers a return to the Learning Phase to restart exploration and adapt to the new network state. This mechanism ensures continued adaptation in dynamic environments.

\subsection{INT-Metadata Collection}
To enable real-time performance monitoring of path segments, our methodology uses INT to collect the relevant metadata at each hop. Each switch along the path appends its instantaneous queue occupancy and dequeueing delay to the packet's telemetry header. 

Upon reaching the domain's sink node $e_k$, each packet is processed to extract and remove the telemetry header. The switch then creates a clone containing only the extracted INT metadata (telemetry report), which is forwarded to a dedicated collector switch connected to the sink. The original packet, now stripped of its telemetry header, continues along its forwarding path, either to the destination host $h_d$ or to a downstream switch of the next domain, depending on the topology.

In the collector switch, the telemetry data is used to derive segment-level performance metrics. Queue length and dequeueing delay values appended at each hop are aggregated to reflect the overall congestion condition of the segment. 

\section{Experimental Evaluation}
\label{sec:evaluation}

\subsection{Setup and Implementation}
\label{sec:setup}
\begin{figure}[t]
    \centering
    \includegraphics[width=0.8\linewidth]{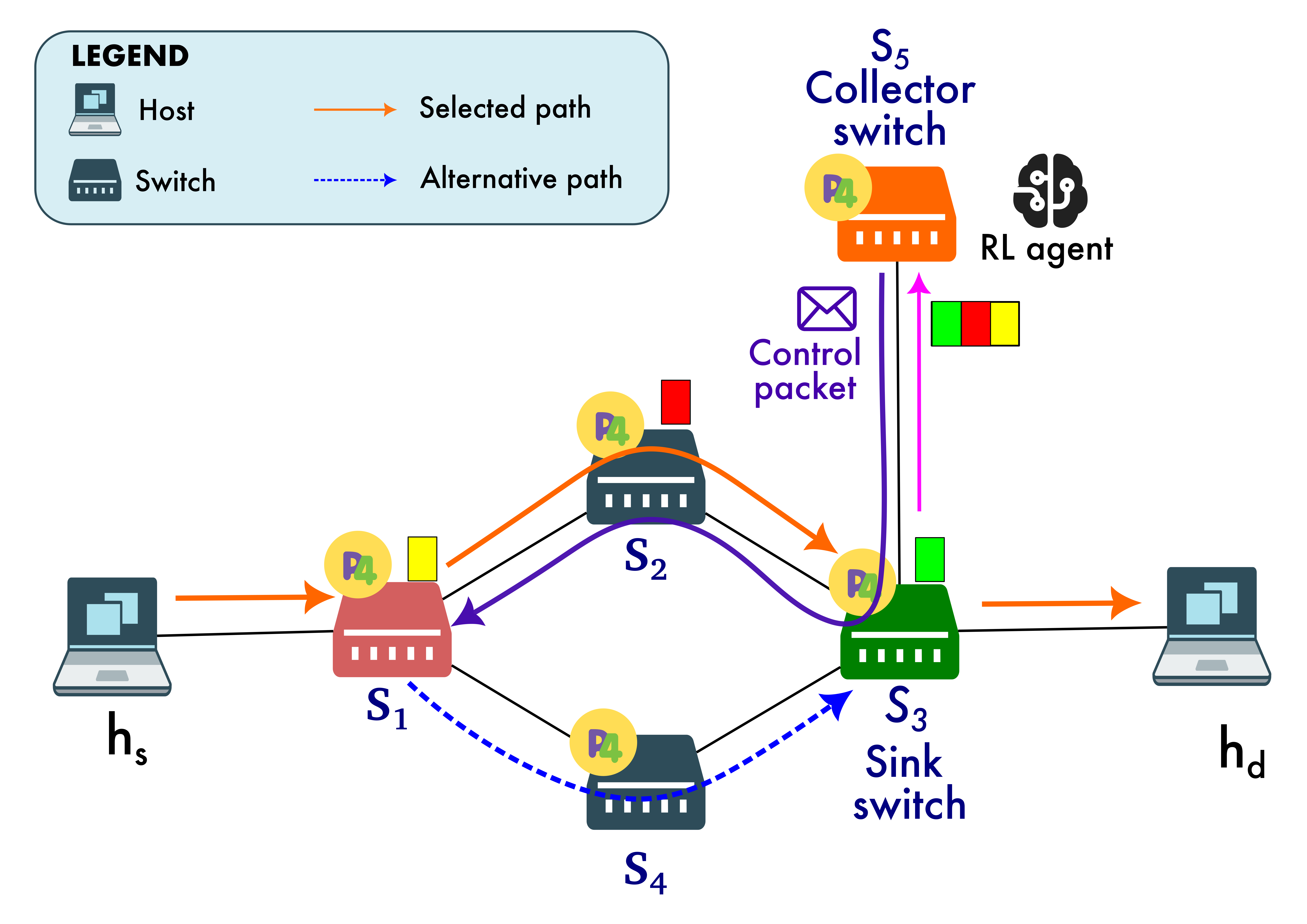}
    \caption{Proof-of-Concept topology with five BMv2 switches and two hosts, including decision node ($S_1$), sink ($S_3$), and collector ($S_5$).}
    \label{fig:PoC}
\end{figure}

We evaluate our IN-RL adaptive path selection methodology through a proof-of-concept implementation in Mininet using five P4-programmable BMv2 switches and two hosts, as shown in Figure \ref{fig:PoC}. The topology represents a single decision domain, where source host $h_s$ transmits traffic to destination host $h_d$ through one of two alternative paths: $p_1: S_1 \rightarrow S_2 \rightarrow S_3$ and $p_2: S_1 \rightarrow S_4 \rightarrow S_3$. Switch $S_1$ acts as the decision node and INT source; switch $S_3$ is the sink node; and switch $S_5$ serves as the collector, hosting the RL agent. 

Implementing RL for path selection under the constraints of the P4 language and BMv2 targets poses several challenges. To address them, key implementation choices were made: path probabilities are stored in registers, simple multiplications are approximated using bitwise shifts, and more complex functions—such as the sigmoid transformation in Eq.~\ref{eq:sigmoid_function} are precomputed and applied via match-action tables. 

To evaluate our implementation, we use iPerf to generate continuous TCP traffic from $h_s$ to $h_d$. This setup allows the RL agent to operate under realistic load conditions and to adapt its path selection based on real-time network state.

\subsection{Results}
\label{sec:results}

We now present the experimental results of our PoC, focusing on how the RL agent responds to real-time congestion. The evaluation examines the agent’s ability to adapt its forwarding decisions based on queue length and dequeueing delay, highlighting how the SLA-based update mechanism enables both stability and responsiveness in traffic steering.

\begin{figure}[t]
    \centering
    \includegraphics[width=\linewidth]{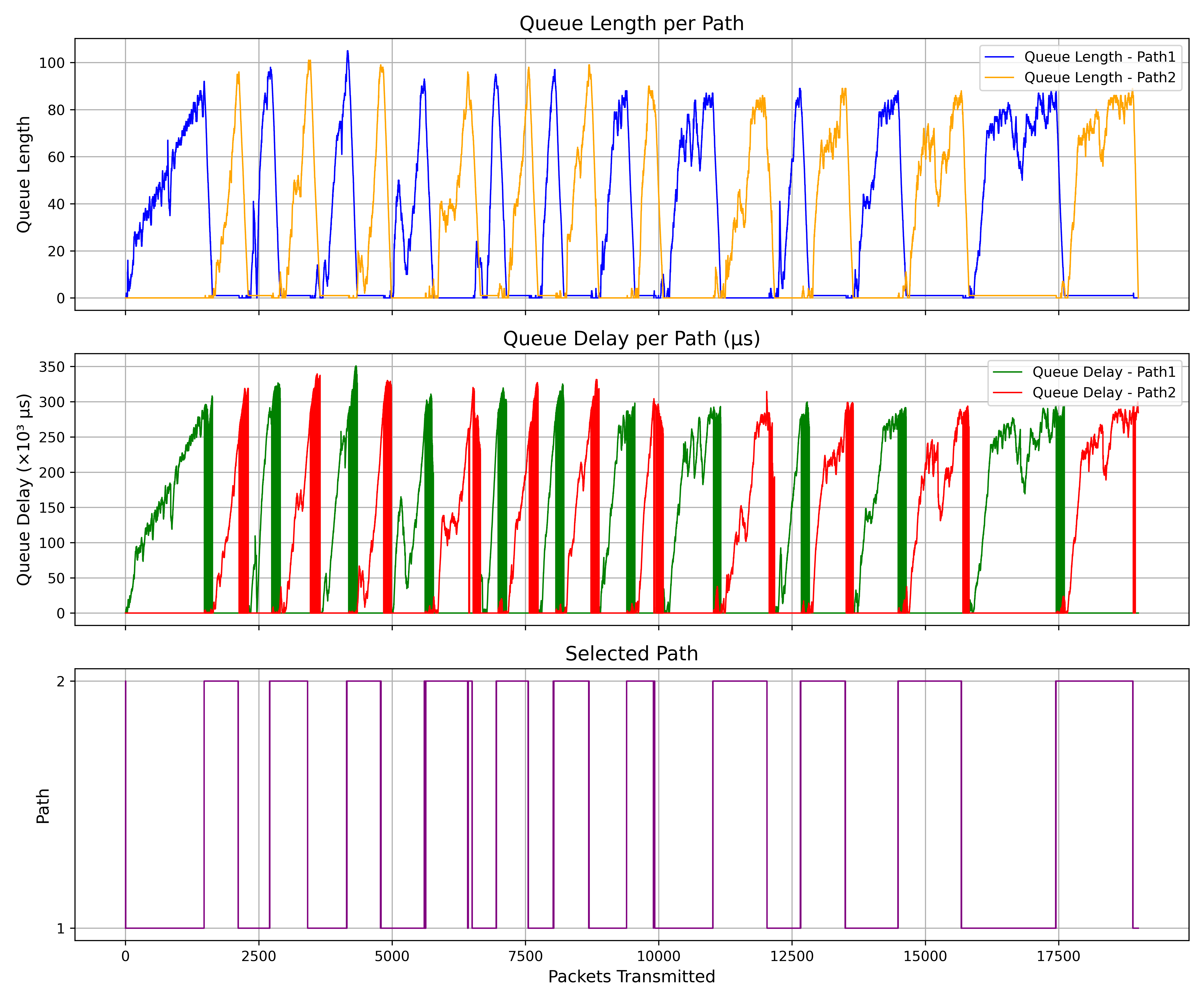}
    \caption{Evolution of queue length, dequeueing delay, and  path selection for $\alpha = 0.5$.}
    \label{fig:combined_metrics}
\end{figure}

Fig.~\ref{fig:combined_metrics} presents the evolution of queue length, dequeueing delay, and selected path over time for a learning rate of $\alpha = 0.5$. This value was chosen based on its favorable convergence behavior, as discussed later in this section.

The top plot shows the queue lengths measured at the collector switch for each candidate path. These values exhibit a buildup when a path is selected and a release when the agent switches to the alternative. The middle plot depicts the corresponding dequeueing delay, which similarly increases with sustained path utilization. Together, these metrics capture congestion trends and provide meaningful input to the agent’s reward computation. The bottom plot displays the selected path over time. This shows that, as congestion builds on the active path—evident from rising queue length and delay—the agent gradually shifts traffic to the alternate path. This demonstrates the reward function's effectiveness in detecting degradation and prompting adaptive redirection.

Moreover, the agent avoids erratic oscillation and instead maintains stable forwarding along the better-performing path until conditions justify a switch. This validates the SLA-based update mechanism’s ability to balance convergence and responsiveness to network dynamics.

\begin{figure}[t]
    \centering
    \includegraphics[width=\linewidth]{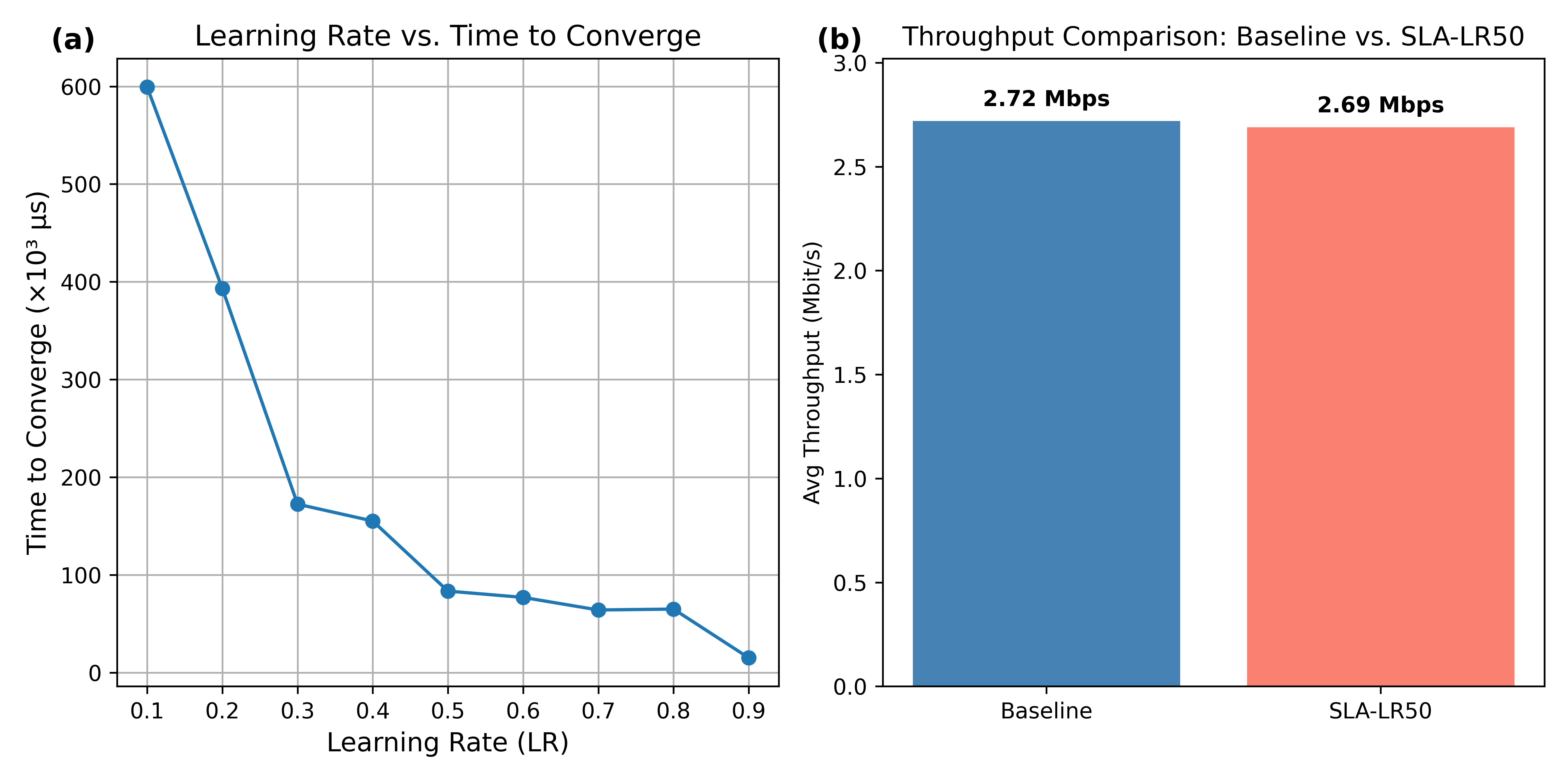}
    \caption{Performance evaluation: (a) Impact of learning rate on convergence time, (b) average throughput comparison between baseline forwarding and SLA-based forwarding with $\alpha=0.5$}
    \label{fig:lr_and_throughput}
\end{figure}

To determine an appropriate learning rate, we evaluated the agent across $\alpha$ values ranging from 0.1 to 1, recording the time required for convergence. Convergence was defined as the moment when one path's selection probability exceeded the threshold of 0.9. As illustrated in Fig.~\ref{fig:lr_and_throughput}(a), increasing the learning rate reduces convergence time by limiting the extent of path exploration, particularly between low and mid-range values. For values beyond $\alpha = 0.5$, the improvement becomes marginal, with convergence times stabilizing below 100 ms. Based on this analysis, we adopt $\alpha = 0.5$ in our experiments as it offers a favorable trade-off between learning speed and stability.

To evaluate the impact in terms of performance, we compared the average throughput achieved in two scenarios: (i) a baseline configuration with standard IPv4 forwarding, and (ii) SLA-based forwarding with $\alpha = 0.5$. Fig.~\ref{fig:lr_and_throughput}(b) reports the results from six iPerf runs per scenario. The observed 1.1\% reduction in throughput reflects the runtime overhead introduced by the SLA-based forwarding logic and telemetry mechanisms. This shows that the impact is minimal, demonstrating that the system maintains near-baseline performance while enabling adaptive, congestion-aware traffic steering.

\section{Conclusions}
\label{sec:conclusions}
This work presented a proof-of-concept implementation of a distributed, in-network reinforcement learning (IN-RL) framework for adaptive path selection in programmable networks. By combining Stochastic Learning Automata (SLA) with real-time telemetry data collected via In-Band Network Telemetry (INT), the proposed system enables local, data-driven forwarding decisions that adapt dynamically to congestion conditions.

The system was evaluated on a Mininet-based testbed using P4-programmable BMv2 switches, demonstrating that queue length and dequeueing delay—collected hop-by-hop—can be effectively used as local indicators of congestion. The RL agent, implemented directly in the data plane, successfully learns to steer traffic away from overloaded paths and maintain stable behavior without oscillation.

Experimental results confirmed the agent’s responsiveness and convergence capabilities, with a learning rate of $\alpha = 0.5$ providing a strong balance between exploration and adaptation speed. This design supports scalability across multiple decision domains and extensibility toward additional performance metrics or objectives.

Future work will explore deployment in hardware-based testbeds, coordination across multiple decision domains, extension of the metric set to include application-level metrics, and evaluation under more dynamic or large-scale network scenarios. 

\section*{Acknowledgment}
This work has been supported by the National Growth Fund through the Dutch 6G Flagship Project "Future Network Services" and by the European Commission H2020 project DESIRE6G (101096466).

\bibliographystyle{IEEEtran}


\end{document}